\documentclass{article}

\usepackage[final]{neurips_2026}
\makeatletter
\renewcommand{\@noticestring}{Workshop on Who Verifies the Agents? Toward Reliable Agent Development, NeurIPS 2026.}
\makeatother

\usepackage[utf8]{inputenc}
\usepackage[T1]{fontenc}
\usepackage{hyperref}
\usepackage{url}
\usepackage{booktabs}
\usepackage{amsmath}
\usepackage{amssymb}
\usepackage{graphicx}
\usepackage{xcolor}

\title{ArrivalBench: Agent-Generated Data Pipelines Are Correct Once
and Wrong Under Time}

\workshoptitle{Who Verifies the Agents?}

\author{Pranay Kothari\\ Department of Computer Science\\ University of Oxford\\ \texttt{pranay.kothari@cs.ox.ac.uk}}

\begin{document}

\maketitle

\begin{abstract}
Benchmarks for agent-generated data work grade a pipeline by executing it once
against a fixed snapshot. We show that such a check cannot discriminate
temporal correctness even where it certifies nearly everything. On 40 tasks
built to isolate temporal behaviour, our reimplementation of single-execution
grading certifies 86--100\% of the pipelines produced by eleven models from
eight organisations, ten of the eleven at or above 97.9\%; re-executing the
same artifacts under adversarial but replayable delivery schedules separates
them across 7.0\% to 79.2\% silent failure, an elevenfold range, though its
shape is two frontier models separating from a cluster rather than a graded
ordering. ArrivalBench verifies what an agent left behind rather than what it
did: the produced pipeline is re-executed after the agent stops, and its final
state must equal a batch recomputation of the complete logical log under every
schedule. Because the oracle recomputes rather than classifies, a wrong answer
and a crash are distinct verdicts, and they are not equally costly: a crash
surfaces at once, while a wrong revenue rollup is found a quarter later. First,
in every arm whose prompt carries no hazard hint, the weakest ordering hazard
scores above the weakest idempotency hazard, so exactly-once is the weaker
axis, though in the two weakest models that margin means less because ordering
itself has collapsed. Second, a prompt can relocate failure rather than remove
it: a hazard warning cuts one model's silent failure from 48.2\% to 10.5\%
while raising its crash rate from 9.0\% to 37.0\%, improving all-in failure
only from 51.0\% to 44.0\%. Of seven intervention arms five repair, one
relocates, and one leaves the model no better than silence. Finally, four
practising engineers given the identical prompt on five of the tasks fail
ordering where the models fail exactly-once; on seven to nine usable cells
that is a dissociation, not a proof, and one model arm inverts on the same
tasks. All eleven unhinted arms were independently re-run: rates move at most
5.9 points and the failing tasks largely recur.
\end{abstract}

\section{Introduction}
\label{sec:intro}

A pipeline that computes yesterday's revenue correctly, once, on a clean batch,
establishes very little. Production data arrives late, arrives twice, arrives
out of order, and arrives again after a crash. Kafka documents at-least-once
delivery in which ``messages are never lost but may be
redelivered''~\citep{kafka_docs}; Debezium warns that consumers ``may likely see
at least some messages duplicated during recovery''~\citep{debezium_docs}; Spark
specifies that ``data later than the threshold will start getting
dropped''~\citep{spark_docs}.

\begin{figure}[t]
\centering
\includegraphics[width=\textwidth]{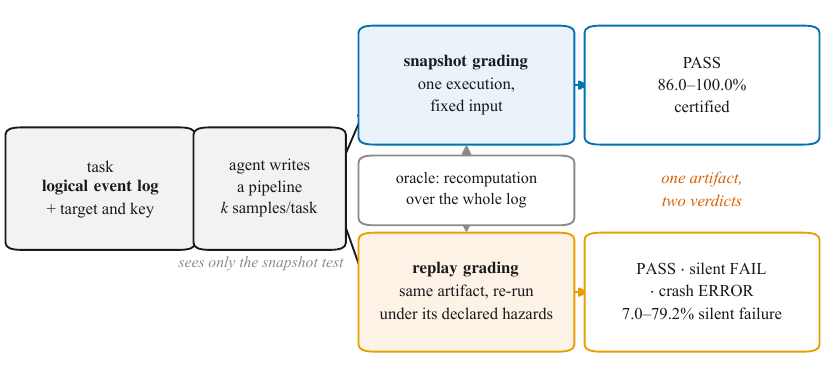}
\caption{Overview. The agent sees only the snapshot test while it writes a
pipeline. The stored artifact is then graded twice: on a fixed input, as the
seven stateful benchmarks we audited do, and under schedules that deliver the
same records late, twice, out of order, after a crash or as a backfill. The
oracle recomputes the target from the complete log, so a wrong table and a
crash are separate verdicts.}
\label{fig:overview}
\end{figure}

Handling these conditions is baseline competence, and it is also a requirement
specifications omit. A production ticket asks for a daily revenue rollup; it
does not ask for tolerance to at-least-once redelivery. The knowledge lives
with the engineer, not in the specification, which is exactly why an evaluation
must probe it deliberately: the prompts these systems receive rarely state
it.

Existing evaluations do not probe it. Of nine benchmarks we audited, seven
produce a stateful artifact for which running again is meaningful, and all
seven grade it by executing once against a fixed snapshot
(Section~\ref{sec:related}). A single execution admits no notion of again, of late, or of out of order, so a
pipeline can pass such a benchmark and corrupt a warehouse on its second day.
Running it a second time on the same data is the cheapest test that would catch
this, and it is the one the benchmarks skip.

We ask an agent to write an incremental pipeline, then verify what it left
behind rather than what it did: the artifact is re-executed, after the agent
has stopped, under adversarial but replayable delivery schedules of the same
logical data. The oracle is deterministic:
\[
\texttt{final\_state(pipeline, schedule)} \;=\;
\texttt{batch\_recompute(complete logical log)}
\]
for every schedule. The right-hand side does not depend on the schedule, so any
discrepancy is a temporal defect and nothing else. The oracle is a
recomputation, not a safe/unsafe classification, so a crash and a silently
wrong answer receive different verdicts. That separation carries the saturation
result (Section~\ref{sec:saturation}), the intervention result
(Section~\ref{sec:conversion}) and the diagnosis of a failed mitigation
(Section~\ref{sec:displacement}).

\paragraph{Contributions.} ArrivalBench is 40 tasks with 42 negative controls
and six documentation-grounded delivery hazards, graded by a replay-invariance
oracle whose verdicts reproduce exactly and whose own verification we report
(Section~\ref{sec:verifying}). With it we show that snapshot evaluation
certifies 86--100\% of pipelines while temporal evaluation separates the same
artifacts across an elevenfold range; that the failures share a signature
across eight organisations, the weakest ordering hazard sitting above the
weakest idempotency hazard in all eleven unhinted arms and violated in none of
the eighteen; and that a prompt can convert silent failure into loud failure
rather than remove it, a distinction visible only because FAIL and ERROR are
separate verdicts. One result implicates us rather than the models: an
architecture prompt of ours specified deduplication against stored history,
said nothing about duplicates arriving together, and left unimproved the model
that followed it most closely. And one tests whether the signature is an artifact of our
own task design, partly clearing it against two of the three: a human anchor
of four practising engineers inverts the signature on five tasks.

\section{Related work}
\label{sec:related}

\paragraph{How data-agent benchmarks grade.} Two of the nine benchmarks we
audited are text-to-SQL~\citep{spider2_2025,beaver2024}, where single execution
is the appropriate design. The remaining
seven~\citep{eltbench2025,kramabench2025,dacode2024,agenticdatabench2026,
cdrbench2026,adebench,dataagentbench} produce a stateful artifact and still
grade it once, by diffing a dumped target against ground truth, scoring a
decomposed reference pipeline, or comparing artifacts under static scoring
modes. Two deserve mention. ADE-Bench~\citep{adebench} is the closest, grading by
snapshots before and after a single agent execution; its variants vary which
database or project, not delivery over time.
DataAgentBench~\citep{dataagentbench} reports Pass@1 over five agent runs, which
estimates variance with byte-identical data, not re-execution under varied
delivery.

\paragraph{Unprompted defaults in code generation.} ArrivalBench belongs to
this line, not outside it. SPARK~\citep{spark2026} establishes the phenomenon:
under a plain prompt, frontier models average ``only 4.63\% safe code in C++'',
and an explicit instruction recovers ``60 or more percentage points in every
language''. Our elicitation result is a domain extension of this, not a
discovery.
SecureForge~\citep{secureforge2026} bounds how far prompting goes: frontier
models ``still produce verifiable vulnerabilities on average 23\% of the time''
even when asked for secure code, and single-character mutations can flip
generated code from secure to vulnerable~\citep{promptfragility2026}. Our
displacement finding is anticipated as a thesis by \citet{sysprompt2026}, who
reports that narrowing instructions to specific signals can degrade
already-capable models by up to 19 percentage points; ours is a measured
mechanism, since the mediating variable is observable in the generated SQL.

What this setting adds is the separation of a wrong answer from a crash.
Safe/unsafe classification cannot make it, because a program that crashes and
one that silently returns a wrong answer carry the same label. This is a
structural property of that grading design and not a criticism of the work;
the word crash simply does not arise under it. Our oracle is a batch
recomputation, so the two are different verdicts, and
Section~\ref{sec:conversion} is what that separation buys.

\paragraph{Temporal correctness as a formal property.}
\citet{lookaheadfreedom2026} reformulates look-ahead bias as ``temporal
non-interference over a time-indexed information lattice''. The vocabulary
overlaps and the properties do not: theirs forbids information flowing
backward in time, ours requires identical final state under any delivery
order. Neither implies the other, since a look-ahead-free pipeline can still
double-count on redelivery, which is most of what we measure. That work
statically verifies hand-written pipelines; ours executes generated ones.
\citet{lookaheadbench2026} shares the vocabulary but measures whether a model
memorised future information. AutoStreamPipe~\citep{autostreampipe2025} scores
generated streaming pipelines by whether they run, which is where we begin.
Stream processors state the guarantees our hazards probe: the Dataflow model
separates event time from processing time and handles late data with
watermarks~\citep{akidau2015dataflow}, and Apache Flink keeps exactly-once
state through consistent snapshots~\citep{carbone2017flink}. Closest in spirit is work that varies the
environment an agent acts in~\citep{gaia2_2026}, which perturbs a live setting
and judges the agent's trajectory. The unit of judgement differs: that verifies
what an agent did, while we verify what it left behind, re-executing the
artifact after the agent has stopped. An agent can act correctly throughout and
still emit a pipeline that corrupts state on its second run.

What is new here is not the elicitation result, a replication of
\citet{spark2026} in a new domain, but the replay-invariance oracle applied
after the agent has stopped, the FAIL/ERROR separation that turns a 37.7-point
improvement into a 7.0-point one, and the human dissociation.
Appendix~\ref{app:comparison} gives one row per audited benchmark.

\section{Method}
\label{sec:method}

\paragraph{Convergence invariant.} Each task defines a logical event log. A
schedule partitions it into delivery batches, possibly duplicating events,
reordering them, withholding and later releasing them, or aborting mid-batch.
The pipeline runs once per batch against a persistent database, and correctness
is the equality above, evaluated after the final batch. This is metamorphic
testing~\citep{metamorphic2018} with an environmental transformation: the
delivery schedule varies, not the input content or the prompt, so logical
information content is identical across schedules by construction. The design borrows from
systems practice rather than benchmark practice: fault injection with a checker
that infers anomalies from observed histories~\citep{elle2020}, and
deterministic simulation testing, where a seeded scheduler makes an adversarial
ordering replayable~\citep{foundationdb2021}. Both test a system an engineer
wrote; we apply the idea to an artifact an agent wrote and then abandoned.

\paragraph{Hazards.} Six hazards are used, each grounded in a verbatim quote
from primary documentation: late arrival, out-of-order change data, duplicate
delivery, crash-and-retry, backfill, which both re-delivers processed events
and releases withheld ones, and additive schema evolution. Two benign controls,
a single in-order batch and a multi-batch in-order stream, separate ordinary
incrementality bugs from temporal ones. Appendix~\ref{app:limitations}
addresses why a watermark that drops late data is graded as a failure here. Five of the six fail silently; one does
not. Schema evolution produced no FAIL verdicts in any arm, because a pipeline
naming a not-yet-existent column raises instead of quietly corrupting. It is
reported separately and never averaged with the other five.

\paragraph{Execution contract.} Agents receive the task, the source and target
schemas, and a contract stating that their script runs once per batch against a
persistent database with no cumulative table of past arrivals. They are told
nothing about any hazard; a word-boundary scan enforces this.

\paragraph{The agent.} Each cell is a loop of at most three turns: the agent
returns one DuckDB script, the harness runs it on the snapshot path only and
hands back a DuckDB error or a typed diff, and the loop ends when the script
passes. No feedback names a hazard and only the final script is graded. The
loop therefore optimises against exactly the check the incumbent metric
applies, so snapshot-pass here is that metric's best case: 79.5\% of cells pass
on the first turn and the loop lifts the per-arm figure by up to 32.5 points
(Appendix~\ref{app:agent}, with the three prompts verbatim).

\paragraph{Admissibility gate.} Every task passes mechanically before
admission: the gold pipeline passes every declared hazard and both benign
controls; each negative control fails exactly where declared while passing
snapshot evaluation, so that it is silently, not visibly, broken; every
declared hazard actually perturbs that task's delivery; and no hazard
vocabulary leaks.
The gate rejected five vacuous negative controls, two over-declared hazard
sets, and one under-powered generator. That first rejection is a reusable
lesson: duplication sensitivity is a property of the task's aggregation
algebra~\citep{dbsp2022} --- under min, max or existence semantics a duplicate
tuple changes nothing --- so negative controls must be chosen per task, not per
archetype.

\paragraph{Metrics and denominators.} A cell is one (task, sample) pair, with
$k=5$ samples per task. Every rate is a mean over tasks of that task's own
rate, never pooled over cells. That is what makes the two-level bootstrap
valid and why reported figures are not round fractions. We report snapshot-pass
as a gating diagnostic; the conditional gap as the primary metric, being
$P(\text{temporal-fail} \mid \text{snapshot-pass})$ over usable cells, that is,
the silent-failure rate among artifacts an existing-style benchmark would have
certified; per-hazard temporal-pass over cells where that hazard itself did not
error; and ERROR-inclusive failure over all cells. The middle two use different
exclusion rules, so a loud failure on one hazard can mask a silent failure on
another in the same cell.

\paragraph{Zero-event rates and determinism.} A bootstrap over an all-zero
sample returns $[0.0, 0.0]$, which reads as certainty and is not, so any rate
at exactly 0\% or 100\% is reported with a rule-of-three one-sided 95\% bound
of approximately $3/n$ over tasks; zero failures over 40 tasks is consistent
with a true rate near 7.5\%. The same (task, schedule, seed, artifact) yields
byte-identical verdicts. The 28{,}350 stored verdicts are 18 arms $\times$ 40
tasks $\times$ $k=5$, each cell graded on a snapshot, two benign controls and
its four or five declared hazards. Of those, 28{,}041 are gradeable; the remaining 309 belong to the 39 cells
where the agent emitted no pipeline, so no artifact exists to re-grade
(Section~\ref{sec:verifying}). All 39 are accounted for: 25 are API refusals
in the three Opus arms, and the other 14 are the truncated cells discussed in
Appendix~\ref{app:limitations}.

\section{Results}
\label{sec:results}

Every table and figure is generated from one machine-readable results file;
no number below is hand-entered.

\subsection{Verifying the oracle}
\label{sec:verifying}

A benchmark that grades by execution is only as good as its grader. All
28{,}041 gradeable verdicts re-grade from stored artifacts with zero
mismatches. Four cells were re-derived by hand with the oracle out of the loop, one a
known failure, so the check confirms the oracle \emph{detects} failure rather
than merely agreeing that passes passed. The admissibility gate rejected work of ours before any
model ran. Two harness defects survived into the runs and were found by audit;
Appendix~\ref{app:limitations} gives their corrections and deltas, the larger
moving one arm's gap by 2.0 points. A third was found by re-running an arm, and
could only abort a run rather than mis-grade one. Three claims from our earlier
pilots did not replicate at 40 tasks and are withdrawn.

\subsection{Snapshot evaluation cannot separate these models}
\label{sec:saturation}

\begin{table}[t]
\centering
\small
\setlength{\tabcolsep}{4pt}
\caption{All unhinted arms, 40 tasks, $k=5$. The left half reports the
incumbent methodology against ours; the right half reports the failure
signature, where \emph{ord.} is the weaker of the two ordering hazards and
\emph{idem.} the weakest of the three idempotency hazards. Open-weights arms
(lower block) were served through a hosted API aggregator. We report current
models only, one per organisation except DeepSeek, whose V4 Flash and V4 Pro
are concurrent tiers of one generation.}
\label{tab:saturation}
\begin{tabular}{llrrlrrr}
\toprule
& & \multicolumn{3}{c}{saturation} & \multicolumn{3}{c}{signature} \\
\cmidrule(lr){3-5}\cmidrule(lr){6-8}
Model & Organisation & snap. & gap & 95\% CI & ord. & idem. & margin (pp) \\
\midrule
Haiku 4.5    & Anthropic & 98.9\%  & 52.2\% & [37.4, 66.9] & 78.7\%  & 51.1\% & $+27.6$ \\
Sonnet 5     & Anthropic & 100.0\% & 48.2\% & [32.8, 63.6] & 98.9\%  & 51.8\% & $+47.1$ \\
Opus 5       & Anthropic & 100.0\% & 16.7\% & [8.5, 26.4]  & 100.0\% & 84.3\% & $+15.7$ \\
GPT-5.6 Sol  & OpenAI    & 100.0\% & 7.0\%  & [1.5, 14.5]  & 100.0\% & 93.0\% & $+7.0$ \\
\addlinespace
Qwen3-Coder-Next  & Alibaba  & 99.2\%  & 73.7\% & [60.6, 85.7] & 48.0\% & 37.0\% & $+11.0$ \\
DeepSeek V4 Flash & DeepSeek & 97.9\%  & 52.4\% & [38.6, 66.2] & 77.6\% & 50.2\% & $+27.4$ \\
DeepSeek V4 Pro   & DeepSeek & 100.0\% & 45.3\% & [31.9, 59.0] & 96.3\% & 54.4\% & $+41.9$ \\
Gemma 4 26B       & Google   & 99.5\%  & 54.7\% & [39.1, 70.4] & 96.2\% & 47.4\% & $+48.8$ \\
MiniMax M3        & MiniMax  & 99.5\%  & 50.2\% & [38.0, 62.6] & 85.1\% & 52.0\% & $+33.1$ \\
Codestral         & Mistral  & 86.0\%  & 79.2\% & [67.2, 89.9] & 34.3\% & 33.4\% & $+0.9$ \\
Kimi K3           & Moonshot & 99.5\%  & 35.2\% & [22.7, 48.8] & 98.7\% & 66.8\% & $+31.9$ \\
\bottomrule
\end{tabular}
\end{table}

\begin{figure}[t]
\centering
\includegraphics[width=0.92\textwidth]{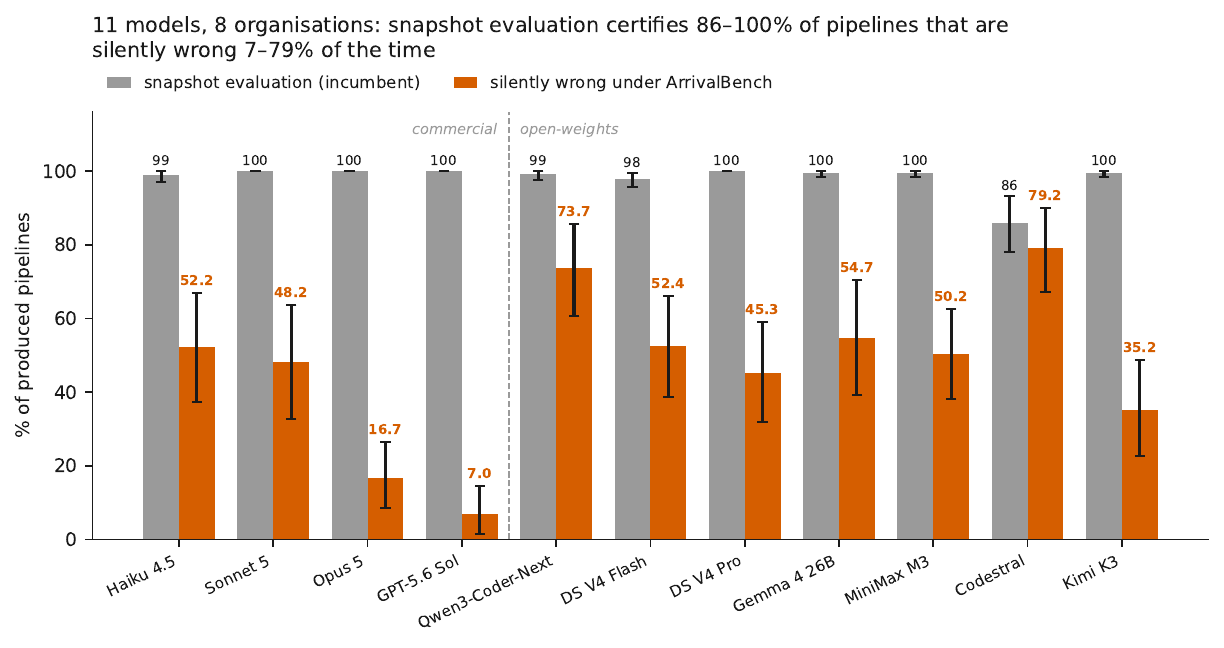}
\caption{Snapshot-pass (grey) against silent-failure rate (orange), eleven
models from eight organisations. Error bars are 95\% bootstrap CIs over tasks
on both series; most snapshot intervals are too tight to resolve.}
\label{fig:saturation}
\end{figure}

Table~\ref{tab:saturation} and Figure~\ref{fig:saturation} state the central
result: snapshot evaluation --- our implementation of the single-execution
grading the seven stateful benchmarks of Section~\ref{sec:related} use, run on
our tasks --- assigns near-identical, near-perfect scores to eleven models
whose temporal correctness differs elevenfold, and does so at its best: our
repair loop optimises against that check (Table~\ref{tab:repairloop}). One arm qualifies the word saturated. Codestral falls to 86.0\% snapshot-pass
and its weakest ordering hazard is 34.3\%, so it fails broadly, not only
temporally. It anchors both ends of the range, so we give the range without it:
over the other ten arms the gap still spans 7.0--73.7\%.

The repair loop does not produce the gap. Among final snapshot passes, the
cells that needed a repair turn fail at least one declared hazard, silently or
by crashing, more often than the cells that passed at once: 69.5\% against
47.7\%, a difference of $+21.9$pp [$+13.0$, $+31.1$]. Those cells sit on harder
tasks. Within the same arm and task, over the 163 strata holding both kinds of
cell, the difference is $+0.2$pp [$-6.6$, $+6.6$], and a linear probability
model with a task random intercept puts it at $+5.0$pp [$+0.5$, $+9.6$].
Repairing against the snapshot test may add a few points of replay failure,
but most of the raw difference is task difficulty. The rates are pooled over
cells with a task-clustered bootstrap, and the three questions of this analysis
were fixed before any intermediate turn was graded.

Every unhinted arm was independently re-run from scratch, a further 2{,}200
cells carrying 17{,}325 verdicts; rates move by at most 5.9 points and the
failing pipelines largely recur (Appendix~\ref{app:limitations}).

The gap conditions on snapshot-pass, not on the benign controls. Recomputing
each gap over only the cells passing both moves eight of the eleven down and
none up, by at most 16.1 points, leaving the range at 7.0--57.6\% on the same
ten arms, so the spread is not an artifact of counting incrementality bugs as
temporal defects (Appendix~\ref{app:limitations}).

We claim no ordering between the two smaller commercial models, each paired
delta being the second-named arm's gap minus the first's: Haiku versus
Sonnet is $-5.7$pp [$-23.7$, $+12.3$], inconclusive, as is Opus versus GPT-5.6
Sol at $-9.7$pp [$-20.7$, $+0.6$]. All four comparisons between the smaller pair and a
frontier model are conclusive, running from $-32.1$pp [$-47.2$, $-16.7$] for
Sonnet against Opus to $-45.6$pp [$-59.7$, $-31.6$] for Haiku against GPT-5.6
Sol. The open-weights arms share one aggregator and carry no
intervention arm, so we make no pairwise capability claim among them. The shape of the result is
therefore that two frontier models separate from a cluster of six whose
intervals each contain the others' point estimates --- Haiku, Sonnet, both
DeepSeek tiers, Gemma and MiniMax --- with Kimi K3 below that cluster and
Qwen3-Coder-Next above it, rather than a graded ordering across eleven.
Against a metric that separates none of them, that is the finding.

\subsection{Exactly-once is the weaker axis in every arm}
\label{sec:signature}

\begin{figure}[t]
\centering
\includegraphics[width=0.82\textwidth]{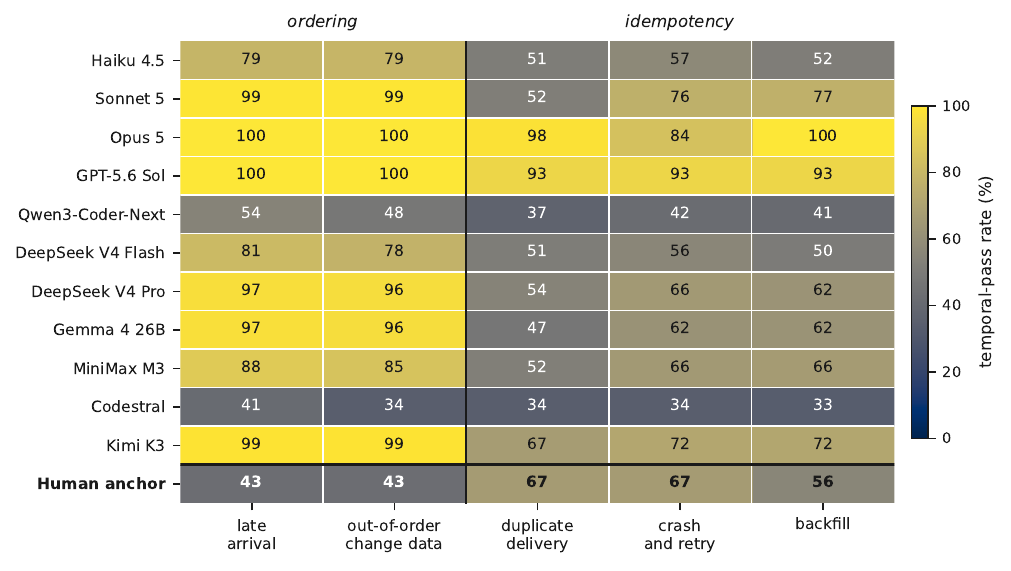}
\caption{Temporal-pass rate per hazard. The rule separates the two ordering
hazards from the three idempotency hazards. Schema evolution is excluded as
the loud hazard (Section~\ref{sec:method}). Human cells are seven to nine,
model cells 128--200. Model rows are means over tasks; the human row is
pooled over cells (Section~\ref{sec:humanarm}).}
\label{fig:signature}
\end{figure}

Stated as a rule, the weakest ordering hazard scores at least as high as the
weakest idempotency hazard. Over all 18 arms it holds strictly in 15, with 3
saturated ties and no violations; over the eleven unhinted arms it is strictly
positive in every one. We report a sign test, with its null, in
Appendix~\ref{app:limitations} and do not lean on it: the weakest-link form is
biased toward the claim by construction, and the arms share 40 tasks, one
engine and one hazard design, so they are not independent draws. The
conservative form, comparing weakest ordering against strongest idempotency,
is what carries the rule: eleven of the eighteen are strictly positive, seven
are saturated ties and none is violated, the ties being six intervention arms
and Opus 5 (Table~\ref{tab:intervention}). We report the weakest-link form
alongside it because that form collapses whenever a single idempotency hazard
saturates. Section~\ref{sec:humanarm} is a small
human anchor pointing the other way, which is a dissociation rather than a
designed control.

Figure~\ref{fig:signature} shows the structure directly, and that it appears
across eight organisations is stronger evidence that the axis is real than any
single rate. Two caveats keep it proportionate.
Codestral's $+0.9$pp is effectively a tie here, though it rises to $+5.3$pp
once the task sets are matched (Section~\ref{sec:limitations}), so the tie is
partly an artifact of the declaration imbalance. And in the two weakest arms the
margin means less, because ordering itself has collapsed, to
34.3\% and 48.0\%: there the rule holds for a different reason than where
ordering is near-perfect. Setting both aside, nine arms carry the
weight, with margins from $+7.0$pp to $+48.8$pp. The two smallest are the frontier ones, both small because idempotency is nearly solved rather than because ordering is weak: GPT-5.6 Sol is flat at 93.0\%, and Opus's margin rests entirely on crash-and-retry, 84.3\% against 98.0\% and 100.0\%. The rule holds at frontier tier on one hazard rather than three. A second account of the same data is that every
gold shares one anti-join staging idiom, which the idempotency hazards probe
and the ordering hazards do not, so the signature could be one idiom failed
often against many idioms failed rarely; Section~\ref{sec:humanarm} bears on
this.

\subsection{A human anchor inverts the signature on five tasks}
\label{sec:humanarm}

The other objection to the rule is that our ordering hazards are simply easier
than our idempotency hazards, in which case we have measured our own generator.
The human arm holds fixed the hazards, the prompt, the leak scan, the harness
and the oracle; the task set is five of the forty and the working conditions
differ, so it is an anchor and not a matched control.

Four practising data engineers wrote pipelines for five tasks under the
identical unhinted prompt, verified by the same leak scan and graded through
the same harness and oracle. They consented, self-declared no prior
knowledge of the hypothesis, worked unpaid and untimed on their own machines,
and were asked not to use an AI coding assistant; neither declaration was
enforced, which on a take-home exercise is close to no control
(Appendix~\ref{app:agent}). Nineteen cells were returned and one task
declined. Their
snapshot-pass rate is 68.4\%, below every model in the study, which we
attribute to an unpaid side task without the repair loop the agents were given.

\textbf{On these five tasks the direction reverses.} Their weakest ordering hazard scores 42.9\%
(3/7) against a weakest idempotency hazard of 55.6\% (5/9), a margin of
$-12.7$pp. These five rates are pooled over cells rather than averaged over tasks,
because per-task cell counts run from one to four here
(Appendix~\ref{app:limitations}).
The inversion holds under all three rules we can apply: $-12.7$pp pooled,
$-10.0$pp averaged over tasks, and $-14.3$pp under the stricter denominator
that drops a cell whenever any declared hazard errored. Restricting every model arm to these five tasks, the rule is strictly positive
in nine of eleven against eleven of eleven over the full suite, with one tie
and one inversion to $-4.5$pp. The two axes come apart: the tasks are harder
on the aggregate gap, but the margins move both ways, six arms up and five
down. The reversal is one-sided, and that is the direct answer to the objection: the
humans are mid-table on idempotency and below every arm but Codestral on
ordering, so four practising engineers do not find our ordering hazards easy.
Against the six-arm cluster their idempotency edge is about 4 points and their
ordering deficit about 46, so the inversion follows mainly from that collapse.
Their margin is more negative than all eleven arms, and than ten of eleven
under the stricter rule; the weakest-link statistic is biased toward a positive
margin by construction, and they invert in spite of it.

The two ordering hazards reach the same value independently, late arrival 3/7 and out-of-order change data
3/7, so the statistic is not pinned by one weak hazard: erasing the inversion
needs one further pass in each of them, not one in total. Two cautions point
the same way. First, that aggregate-gap difficulty is substantial: Opus scores 16.7\%
over 40 tasks and 32.0\% over these five. Second, on the aggregate gap they are not better, scoring 57.1\% against a
model range of 0.0--68.0\% on these tasks. Seven to nine cells over five tasks
cannot separate participant from task effects, and our
registration fixes $n=3$--5 as an anchor; we compute no interval and make no
significance claim. The anchor also bears on the staging-idiom account of
Section~\ref{sec:signature}: the participants wrote their own staging step, so
that idiom was not shared with them, and they still failed the diverse-code
axis rather than the shared-idiom one, the opposite of what that account
predicts. Five tasks give the diverse axis little room to vary, so this
constrains the account rather than eliminating it.

\subsection{Where prompting converts silent failure into loud failure}
\label{sec:conversion}

The intervention arms cover three tiers, Haiku 4.5 as small, Sonnet 5 as mid,
and Opus 5 with GPT-5.6 Sol as frontier. A single paragraph warning that data
may arrive late, twice, out of order, or be replayed drives the measured gap to
zero at frontier tier in both families:
Opus 5 from 16.7\% to 0.0\% (95\% upper bound 7.5\%; paired $-16.7$pp
[$-26.0$, $-8.7$]) and GPT-5.6 Sol from 7.0\% to 0.0\% (upper bound 7.5\%;
paired $-7.0$pp [$-14.5$, $-1.5$]). Hinted GPT-5.6 Sol returned 1{,}575 PASS verdicts
out of 1{,}575, consistent under the same bound with a true failure rate up to
7.5\%. Neither zero is produced by the masking noted in
Section~\ref{sec:method}: recomputing both with schema evolution dropped from
the hazard set leaves each at 0.0\%, because neither model ever crashed on
that hazard. The same recomputation moves ten of the eighteen arms, never
by more than 3.3 points and never downward, so masking makes the reported
figures conservative rather than flattering. Below frontier tier the same paragraph leaves a small model silently
wrong 35.1\% [22.8, 47.8] of the time. No existing evaluation issues such a warning, so the
unwarned figure is the one describing an agent given only the task, which is
what current benchmarks measure and what a bare ticket supplies. Deployment
sits between the two arms, since a system prompt or a house coding standard can
function as a standing warning.

The headline improvement at mid tier is not what it appears. Hazard hinting
cuts Sonnet's silent failure rate from 48.2\% to 10.5\%. That marginal
difference is 37.7pp; the comparison of record is the paired estimate, which
resamples tasks carrying both arms' samples, and it gives $-33.4$pp
[$-47.3$, $-20.1$]. The same hint raises the crash rate from 9.0\% to 37.0\%, a
paired $+28.0$pp [$+17.5$, $+39.0$] and slightly more than a fourfold
increase. That also means the two arms' denominators are not composed of the
same cells, which is why we co-report all-in failure; restricted to the 30
tasks usable in every Anthropic arm, a stricter intersection than this pair
alone requires, it reads 36.0\% against 12.6\%. The contrast attenuates from
37.7 to 23.4 points and survives, so the conversion is not an artifact of
which cells survived. All-in failure therefore improves only from
51.0\% to 44.0\%. The mechanism is visible in the generated code: the hint raises the
guard rate to 91.5\%, those guards are predominantly \texttt{PRIMARY KEY}
declarations, and the constraints then fire, producing 71 constraint violations
against 17 unhinted. The small model's architecture arm is the opposite case. There all-in failure
equals the error rate exactly at 32.0\%, every sample that ran at all being
temporally correct, and the error rate barely moved, 30.5\% to 32.0\%. All-in
failure fell from 67.5\% to 32.0\%, the largest such improvement in the study
and a repair rather than a conversion. Both are single-paragraph prompt
additions with opposite outcomes on the same metric, which is the point: only a
grader that distinguishes a wrong answer from a crash tells them apart, and a classifier
labelling code safe or unsafe records both as failure.

\subsection{A partial specification can displace a model's own defences}
\label{sec:displacement}

A third registered arm names a stage-then-derive pipeline shape and mentions no
hazard. It reaches 0.0\% at the small and frontier tiers (one-sided 95\% bound
$\leq 7.5\%$ in each) but 54.4\% at mid
tier. The small tier's zero is the one figure in the paper that the masking
moves: dropping schema evolution puts it at 2.6\%, a single failing cell whose
schema-evolution crash removed it from the denominator. The frontier zero does
not move. This is the only condition in the study where a prompt fails to improve on
silence. The arm itself is stable: it was re-run twice more because the result was
surprising, an asymmetric stopping rule we did not apply to unsurprising arms,
and the three runs give 53.2\%, 54.4\% and 55.5\% (Table~\ref{tab:replication}
covers the unhinted arms only). What is not
established is the comparison against silence. Paired against the unhinted arm
the difference is $+6.5$pp with a 95\% interval of [$-1.6$, $+16.8$], which
crosses zero and is therefore inconclusive under our pre-registered rule, and
pooling every run of both arms moves it only to $+7.7$pp [$-0.3$, $+17.4$]. What is established is the
non-uniform transfer rather than a degradation: the same prompt is conclusively
beneficial for the small model ($-50.9$pp [$-65.9$, $-35.6$]) and drives the
frontier model to zero, while for this model it buys nothing. Every one of that
model's failures under this prompt is on duplicate delivery, and the cause is
in the SQL (Figure~\ref{fig:worked}): it deduplicates against stored history
but not within the arriving batch, so when one batch carries the same event
twice, neither copy is in history yet and both land. The guard is not absent;
it is correct against the wrong scope. Our prompt is responsible, since it asks for ``inserting a given
\texttt{event\_id} only if it is not already present'', which specifies
deduplication against history and is silent on duplicates arriving together. We
wrote it believing it complete, and the model that followed the instruction
most literally was the one it harmed.

\begin{figure}[ht]
\centering
\small
\rule{\textwidth}{0.8pt}\\[3pt]
\begin{minipage}[t]{0.55\textwidth}
\footnotesize
\begin{verbatim}
INSERT INTO sales_history
SELECT b.event_id, b.ts_event,
       b.store_id, b.amount
FROM sales_batch b
WHERE NOT EXISTS (
  SELECT 1 FROM sales_history h
  WHERE h.event_id = b.event_id);
\end{verbatim}
\end{minipage}\hfill
\begin{minipage}[t]{0.42\textwidth}
\footnotesize
\vspace{4pt}
\begin{tabular}{@{}lrr@{}}
\toprule
one logical log, delivered as & revenue & rows \\
\midrule
a single in-order batch & 425.00 & 3 \\
redelivered in a later batch & 425.00 & 3 \\
\textbf{redelivered within one batch} & \textbf{675.00} & \textbf{4} \\
\bottomrule
\end{tabular}
\end{minipage}\\[3pt]
\rule{\textwidth}{0.8pt}
\caption{Staging step of a generated pipeline that passes snapshot evaluation
and is graded FAIL on duplicate delivery. The guard is correct against history,
so a later redelivery is absorbed and a within-batch one is not: revenue is
overstated by 59\% and no error is raised.}
\label{fig:worked}
\end{figure}

Within that model's own arm, with model, prompt and task set held fixed, its
200 cells (40 tasks $\times$ $k=5$) split on whether the generated SQL retained
a within-batch guard. The rates in this section alone are pooled over cells
rather than averaged over tasks as Section~\ref{sec:method} otherwise
requires, because guard retention varies between samples of the same task and
so is not a task-level property; the per-arm sign test below is unaffected,
being one count per arm.
The 67 guarded cells never fail duplicate delivery silently (0 FAIL, one-sided
95\% bound $\leq 4.5\%$), but 40.3\% of them \emph{crash} on it, the
constraint firing on the redelivered row. The 133 unguarded cells never crash
and fail silently 72.2\% of the time. \textbf{The guard does not prevent
failure; it relocates failure}, which is the trade
Section~\ref{sec:conversion} reports for one arm, here visible between
individual pipelines under a single prompt.

\begin{figure}[t]
\centering
\includegraphics[width=\textwidth]{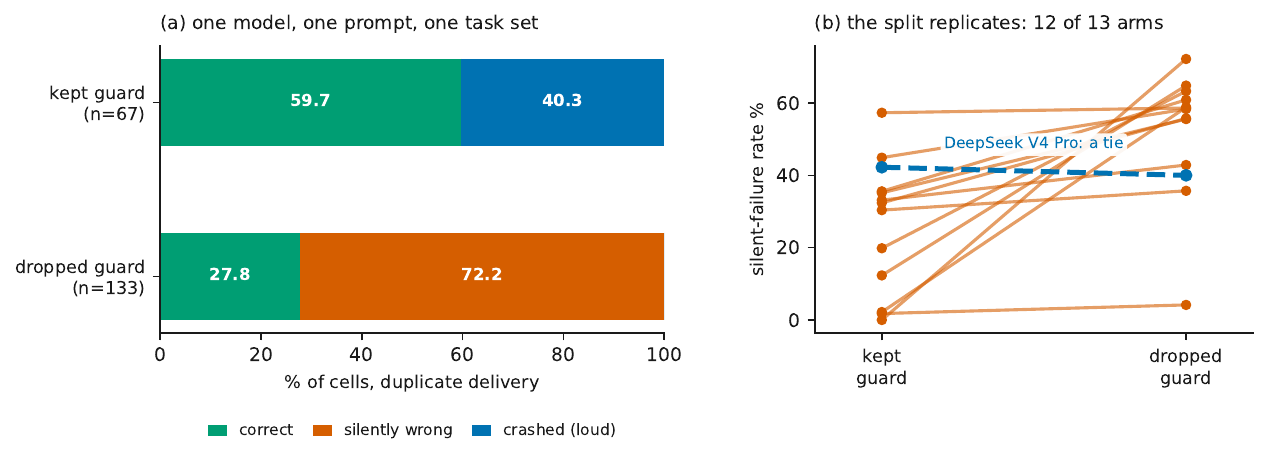}
\caption{(a) Within one arm, cells that kept a within-batch guard crash on
duplicate delivery (40.3\%) and never fail silently; cells that dropped it never
crash and fail silently (72.2\%). The guard relocates failure rather than
removing it. (b) The same
test run independently in each arm: dropping the guard raises silent failure in
12 of 13; the exception, DeepSeek V4 Pro (blue, dashed), is a tie rather
than a reversal.}
\label{fig:displacement}
\end{figure}

\textbf{Guard retention, not capability, is what the within-arm split
isolates.} Running the same cell-level test independently inside every arm
with at least ten cells in both groups,
guarded pipelines silently fail less often than unguarded ones in 12 of 13
arms (Figure~\ref{fig:displacement}b; sign test $p = 1.7\times10^{-3}$, same
independence caveat), and crash more often in 9 of 13.
The single exception is a tie rather than a reversal. The mechanism is
exploratory, identified after the outcomes were known, and is not a controlled
test (Appendix~\ref{app:limitations}).

\section{Discussion and limitations}
\label{sec:limitations}

Across the seven intervention arms, all commercial, what a prompt does varies
and the gap alone does not reveal which: five lower all-in failure with the
error rate flat or falling, one converts silent failure into crashes, and one
leaves the arm worse than unprompted. That is why
the all-in figure is the one to read beside the gap; error composition also
differs by arm (Appendix~\ref{app:limitations}).

Six limitations bound the results. The 40 tasks, their golds and their 42
negative controls were written by one author, and the gate checks conformance
to those declarations rather than their correctness; a validation pass in the
style of \citet{eltbenchverified2026} is future work. Everything runs on
one engine and one dialect, and the signature could in part be a property of
DuckDB SQL. Every gold shares one anti-join staging idiom, which the
idempotency hazards probe and the ordering hazards do not, so the idempotency
half of the signature generalises less far. The two classes are also declared by
different task counts, so their means are taken over slightly different
subsets; recomputed over the 35 declaring all five, the rule stays strictly
positive in eleven of eleven (Appendix~\ref{app:limitations}). Schema evolution is declared by 2 of 40, so the
loud-failure result and the masking analysis describe a twentieth of the
suite. The human anchor is four people and nineteen cells, and it
is our only leverage against the three accounts on which the signature would be
an artifact of our design rather than a property of the models: hazard
difficulty, the shared staging idiom, and the dialect. It is too small to rule
any of them out, so all three remain open. Appendix~\ref{app:limitations} gives these in full.

\section{Conclusion}

Existing evaluation of agent-generated data pipelines asks whether the artifact
is right once. On tasks built to isolate temporal behaviour that question is
uninformative: ten of eleven models answer it at 97.9\% or better. Asking
whether it is right \emph{again} separates the same artifacts across an
elevenfold range, and in every unhinted arm the failures concentrate on
exactly-once semantics rather than event-time ordering. A small human anchor inverts that ordering on the five tasks it
covers, and one model arm with it, so the anchor bounds the task-design and
dialect accounts rather than settling them. Two consequences follow: re-run a pipeline against the same
data before trusting it, and treat a prompt as a specification that is
incomplete. Whether a constraint expressing an invariant beats a query that
assumes it is left open; settling that needs no model calls, only a uniform
template baseline keyed on each task's declared primary key. An artifact can
pass every check that reads it once and still be wrong the second time it
runs.

\section*{Ethics statement}

\paragraph{Human participants.} The human anchor of
Section~\ref{sec:humanarm} involved four practising data engineers who wrote
pipelines from the unhinted prompt given to the models. Participation was
voluntary, unpaid and by invitation among the author's professional contacts.
The packet carried a consent page, and participants could decline any task;
one task was declined. No personal or demographic data was collected, and we
report only pass rates pooled over cells. The exercise carried no foreseeable
risk beyond ordinary professional work. It was not reviewed by an institutional
board, and we report it as a check on a task-design reading, not as a
comparison of humans and models.

\paragraph{Use and misuse.} The benchmark measures whether generated pipelines
corrupt data under replay, and its foreseeable use is defensive. Its event logs
are synthetic, every model was queried through its provider's hosted API, and
no component targets a deployed system. The remaining risk is that the
benchmark is read as a certificate: passing a finite set of schedules is not a
proof of correctness under every delivery history.

\clearpage
\bibliographystyle{plainnat}
\bibliography{refs}

\appendix

\section{How the audited benchmarks grade}
\label{app:comparison}

Table~\ref{tab:comparison} records the grading mechanism of every benchmark in
our audit. Each row was read at PDF or harness level rather than from an
abstract, and the distinction that matters is in the last two columns: whether
the produced artifact is executed more than once under different delivery of
the same logical data, and whether a wrong answer and a crash receive different
verdicts.

Seven of the nine produce a stateful pipeline. The other two are text-to-SQL,
where the artifact is a single \texttt{SELECT} with no persistent state. Grading such an artifact once is inherent to its
type, not an evaluative choice, so they are context for how the field measures
data work and not instances of the gap. The claim bites for the seven whose
artifact is a stateful pipeline: there, re-running is a meaningful operation
and grading exactly once leaves temporal correctness unobserved.

Our audit recorded how each benchmark grades, not what feedback its agent
receives while generating, so we cannot say from our own records how many give
iterative feedback against their own grader as our loop does. The comparison in
this paper does not rest on it: without the loop, snapshot-pass would span
64.0--100.0\% rather than 86.0--100.0\% (Table~\ref{tab:repairloop}), which
still does not resolve these models the way temporal evaluation does, across
7.0--79.2\%, and does not order them the same way.

We do not claim these benchmarks are wrong. Each measures what it set out to
measure, and several are more demanding than ours on axes we do not test at
all, including multi-source discovery, long-horizon tool use and realistic
repository context. The claim is narrower and is the one the last two columns
support: none of them re-executes the artifact under varied delivery, so none
of them can observe the failure mode this paper reports.

\begin{table}[h]
\centering
\small
\caption{How the nine audited benchmarks grade, with Gaia2 and this work for
contrast. \emph{Exec.} is how many times
the produced artifact is run for grading; \emph{FAIL vs ERROR} is whether a
silently wrong result and a crash get different verdicts. Rows marked
$\dagger$ are text-to-SQL: a query holds no state, so single execution follows
from the artifact type rather than from a choice. Gaia2 grades a trajectory,
not an artifact.}
\label{tab:comparison}
\footnotesize
\setlength{\tabcolsep}{5pt}
\begin{tabular}{llccc}
\toprule
Benchmark & Grading & Exec. & Varied & FAIL vs \\
          &         &       & delivery & ERROR \\
\midrule
Spider 2.0$^\dagger$~\citep{spider2_2025}     & execution match & 1 & --- & --- \\
BEAVER$^\dagger$~\citep{beaver2024}           & execution match & 1 & --- & --- \\
\addlinespace
ELT-Bench~\citep{eltbench2025}                & CSV diff        & 1 & --- & --- \\
KramaBench~\citep{kramabench2025}             & sub-task scores & 1 & --- & --- \\
DA-Code~\citep{dacode2024}                    & output match    & 1 & --- & --- \\
AgenticDataBench~\citep{agenticdatabench2026} & static modes    & 1 & --- & --- \\
CDR-Bench~\citep{cdrbench2026}                & final output    & 1 & --- & --- \\
ADE-Bench~\citep{adebench}                    & snapshot diff   & 1 & --- & --- \\
DataAgentBench~\citep{dataagentbench}         & Pass@1, 5 runs  & 1 & --- & --- \\
\addlinespace
Gaia2~\citep{gaia2_2026}                      & action verifier & n/a & n/a & --- \\
\midrule
\textbf{ArrivalBench} & batch recomputation & \textbf{7--8} &
$\checkmark$ & $\checkmark$ \\
\bottomrule
\end{tabular}
\end{table}

Gaia2 is the closest in spirit and the row that most repays reading. It
perturbs a live environment and evaluates the agent's trajectory with an
action-level verifier, so the unit of judgement is what the agent did.
DataAgentBench is the one most easily mistaken for ours: its Pass@1 over five
runs resamples the \emph{agent} against byte-identical data, which estimates
generation variance rather than re-executing one artifact under varied
delivery. Both are answering a different question, not answering ours badly.

The \emph{Executions} figure for our own row is 7--8 per cell: a snapshot, two
benign controls, and the four or five hazards the task declares.

\section{Full limitations and disclosures}
\label{app:limitations}

\paragraph{Scope of the evaluation.} There are 40 tasks, 30 of them LLM-drafted
or templated, all authored by one person without independent review;
\citet{howmanytasks2026} finds that the task fraction needed for reliable
pairwise conclusions varies sharply by benchmark, so we rely on reported
intervals and not on a task-count argument, and a validation pass in the
style of \citet{eltbenchverified2026} is future work. Everything runs on DuckDB
and SQL, and the loud-failure result for schema evolution is a name-binding
artifact of this engine. We use $k=5$, with run 2 reported as replication
and not pooled, and zero-gap arms carry rule-of-three bounds, not demonstrated
zeros. Model scope is eleven models from eight organisations, all via hosted APIs;
the seven open-weights arms were served through one third-party aggregator,
and intervention arms exist only for the commercial models. The
task suite is under active extension; the harness, the tasks with their
negative controls, the pre-registration and the verdict-level results will be
released with the archival version of this work.

\paragraph{The two harness defects in full.} The
response parser required a generated script to open with a statement keyword,
so four cells whose script opened with a comment line were recorded as
producing no pipeline and excluded as model failures. All four fell in the
MiniMax M3 arm. The model's own response was still on disk; re-extracting and
grading it moved that arm's gap from 48.2\% to 50.2\% and its error rate from
18.5\% to 16.5\%. The defect
suppressed real silent failures, and an audit of every arm found it in those
four cells and nowhere else. Separately, a fixed 8{,}000-token output cap was
exhausted by reasoning models on our longest tasks, so fourteen cells across
two arms emitted zero characters after spending the entire budget and were
excluded; the cap applied uniformly to every arm but is not neutral, since it
drops the hardest tasks for the models that reason longest. We re-ran exactly
those cells at a doubled cap. Thirteen of the fourteen then produced a
pipeline, confirming the truncation diagnosis, and the effect is small and not
systematic in direction: DeepSeek V4 Pro's gap moves from 45.3\% to 46.5\%
while MiniMax M3's moves from 50.2\% to 48.9\%. Both shifts sit well inside the
arms' reported intervals and both asymmetry margins remain strongly positive,
moving from $+41.9$ to $+40.2$pp and from $+33.1$ to $+32.6$pp. Headline
numbers retain the uniform 8{,}000-token condition, because giving two arms
more generation budget than the other nine received would not be a
like-for-like comparison, and no conclusion in this paper depends on which
condition is used.

\paragraph{A third defect, found by re-running an arm.} Re-executing
Codestral aborted 18 tasks in. The oracle sorts twice when it builds a diff,
once over rows and once over key tuples, and only the row sort was guarded
against NULLs; a NULL key raised \texttt{TypeError} and killed the run. The
same defect class had already been fixed twice elsewhere, and the regression
test written the second time exercised NULLs in value columns but never in a
key. No verdict was ever wrong, because the failure aborts grading rather than
mis-grading it: every completed run is unaffected, and the verdicts
re-grade byte-identically either side of the correction. A regression test now covers a NULL in a key column, and a
second pins the guarded sort against the unguarded one wherever the latter
worked, so the correction cannot reorder stored diffs. No cell in the corpus
exercises the repaired path, so its behaviour rests on those tests rather than
on the data. Auditing did not surface this defect; re-executing an arm did.

\paragraph{How the re-grade was checked.} The zero-mismatch claim in
Section~\ref{sec:verifying} is one sweep over all eighteen arms, run after the
correction above with no other job touching the artifact directories: 28{,}041
gradeable verdicts reproduce and the 39 cells holding no stored pipeline are
skipped, 25 Opus refusals and 14 truncated, which is the split
Section~\ref{sec:method} reports. Isolation is part of the procedure rather
than a courtesy. An earlier attempt overlapped a live run that was rewriting
the artifacts underneath it and reported 305 spurious mismatches; separately,
enumerating cells by globbing the artifact directory had once matched
same-named files left by a 10-task pilot, so the script takes its cell list
from the verdict file instead.

\paragraph{The registered open-weights roster.} We pre-registered one model
from each of six named organisations. Three never produced an arm, each
abandoned on provider latency or a hang, and a fourth organisation's model was
replaced by its successor; three organisations we had not registered were
substituted to preserve the organisation count. Each substitute was committed
to and run to completion before the next was chosen, so no model was selected
after seeing its results, but the set is not the registered set. Our
registration also under-predicted the severity: we expected most open-weights
gaps in a 15--45\% band and 7 of 8 arms exceeded it, and we expected at least
one thin gap denominator, which did not occur, the minimum being 35 of 40
tasks. Where an organisation ships successive versions we report one current
model per organisation, the exception being DeepSeek's two concurrent V4 tiers;
a superseded model we also ran, Qwen3-Coder 480B, gave
a gap of 72.0\%, snapshot-pass of 100.0\% and a margin of $+4.6$pp, consistent
with the pattern in the main tables, and is retained in our results file.

\paragraph{The bootstrap.} Every interval in the paper is a percentile
bootstrap over 10{,}000 resamples at the 95\% level. The resampling unit is the
task, drawn with replacement to the original task count; a resampled task
carries all $k$ of its samples, which is what the mean-over-tasks rule in
Section~\ref{sec:method} makes valid, and samples are never resampled
independently. Paired comparisons resample the tasks usable in both arms and
take both arms' per-task rates from the same draw, so shared task difficulty
cancels; tasks usable in only one arm are excluded from the pair rather than
imputed. Intervals are not bias-corrected or accelerated, and a rate at exactly
0\% or 100\% is reported with a rule-of-three bound instead of a bootstrap
interval, because resampling an all-zero sample returns $[0,0]$.

\paragraph{Error composition.} Error rates are not comparable across models.
Across the eleven unhinted arms, one arm's errors are entirely API refusals,
seven arms' are entirely SQL execution failures, two mix those with the agent
emitting no pipeline, and GPT-5.6 Sol produced no error of any kind, a zero
carrying the same rule-of-three bound of 7.5\% as the others. The refusals are
orthogonal to the gap: restricting to tasks that model never refused moves its
gap from 16.7\% to 16.5\%.

\paragraph{Provenance.} Corroborating evidence exists that the harness could
not fabricate, since refusals carried by an API stop reason appear in one
model's arms only and output-length profiles separate the arms, but this is
corroboration, not proof, and cost cannot serve as evidence because it
is computed from the requested model's pricing. The open-weights arms are
stronger on this axis: the aggregator's per-request served-model identifier is
present on 1{,}186 of 1{,}400 responses and matches the requested model in
every one of them. It is absent for one arm that ran before we persisted the
field, and for fourteen records that a cell-level re-run superseded without
replacing the original entry.

\paragraph{Metric interaction and verification coverage.} A loud failure on one
hazard can mask a silent failure on another in the same cell, as noted in
Section~\ref{sec:method}. Recomputing every arm's gap with schema evolution
dropped from the hazard set moves ten of the eighteen and always upward, by at
most 3.3 points; the largest absolute consequence is the small model's
architecture arm, whose 0.0\% becomes 2.6\%
(Section~\ref{sec:displacement}), and no other zero moves at all. The by-hand re-derivation covered four
cells and did not implement crash-and-retry semantics, so it verified the
non-crash path for that hazard.

\paragraph{Benign controls.} The two benign schedules are not in the gap's
hazard set, so the gap does not condition on them. Over snapshot-passing
cells, averaged over tasks, the single in-order batch is passed by 100.0\% in
all eleven unhinted arms; the multi-batch in-order stream is passed by 100.0\%
in Opus, GPT-5.6 Sol and Kimi K3, between 89.9\% and 99.0\% in six further
arms, and by 68.2\% in Qwen3-Coder-Next and 60.8\% in Codestral. Restricting
each gap to the cells passing both moves eight of the eleven downward and none
upward, leaving those same three unchanged; the largest moves are 16.1, 9.6
and 7.3 points for Qwen3-Coder-Next, Codestral and MiniMax M3. The two arms with the weakest benign performance are
also the two set aside in Section~\ref{sec:signature}.

\paragraph{Replication in full.} Table~\ref{tab:replication} gives the full
comparison. Every unhinted arm was re-run from scratch with fresh sampling. Task-level agreement matters more than the stability of
the rate: for two arms, no task that failed in run 1 passed in run 2, and per-hazard values reproduce within a few points on independently
sampled pipelines. Gemma moved furthest, by 5.9 points, and downward; Opus
moved by 4.9 with each run's interval containing the other's point estimate,
being [8.5, 26.4] and [12.1, 32.5]. Codestral, whose rate most invites doubt,
moved least of the eleven, by $-0.1$ points; Qwen3-Coder-Next, the other high
arm, moved $+1.9$, the median movement of the table. Two rows would round
differently if the difference were taken before rounding, DeepSeek V4 Pro to
$+2.5$ and Kimi K3 to $-1.5$; Kimi's run-1 gap is exactly 35.25\%, a tie
resolved half-to-even so that Table~\ref{tab:saturation} and
Table~\ref{tab:replication} print the same rate for it.

\begin{table}[t]
\centering
\caption{Independent replication of all eleven unhinted arms. Rates move; the
identity of the failing pipelines largely does not. $\Delta$ is the difference
of the two rates as printed, so every row reconciles on the page.}
\label{tab:replication}
\begin{tabular}{lrrrl}
\toprule
Arm & run 1 & run 2 & $\Delta$ & failing tasks recurring \\
\midrule
Haiku 4.5   & 52.2\% & 54.1\% & $+1.9$pp & 22 of 22 \\
Sonnet 5    & 48.2\% & 44.4\% & $-3.8$pp & 19 of 20 \\
Opus 5      & 16.7\% & 21.6\% & $+4.9$pp & 12 of 14 \\
GPT-5.6 Sol & 7.0\%  & 8.5\%  & $+1.5$pp & 5 of 5 \\
\addlinespace
Codestral        & 79.2\% & 79.1\% & $-0.1$pp & 30 of 33 \\
Qwen3-Coder-Next & 73.7\% & 75.6\% & $+1.9$pp & 29 of 30 \\
Gemma 4 26B      & 54.7\% & 48.8\% & $-5.9$pp & 18 of 21 \\
DeepSeek V4 Flash & 52.4\% & 52.8\% & $+0.4$pp & 22 of 25 \\
DeepSeek V4 Pro   & 45.3\% & 47.7\% & $+2.4$pp & 20 of 25 \\
MiniMax M3        & 50.2\% & 49.3\% & $-0.9$pp & 23 of 29 \\
Kimi K3           & 35.2\% & 33.8\% & $-1.4$pp & 13 of 19 \\
\bottomrule
\end{tabular}
\end{table}

\paragraph{Idiom diversity across the gold pipelines.} A 40-task suite built
by one author invites the question of whether it is one task forty times. It is
not, but one idiom is shared. Every gold stages arrivals with an anti-join on
the event key, which the execution contract effectively forces by providing no
cumulative table of past arrivals, so the idempotency hazards probe a single
staging idiom across the whole suite rather than several. Beyond that step the
golds differ: eleven distinct signatures over the features we tracked, with 22
of 40 using a window function, 13 requiring \texttt{ROW\_NUMBER}
deduplication, 8 rebuilding by delete-and-reinsert and 30 recomputing the
target outright, at lengths from 25 to 53 lines. The shared staging step is a
real limitation on how far the idempotency result generalises, and it is also
what makes Figure~\ref{fig:worked} representative rather than anecdotal.

\paragraph{The signature on the seven intervention arms.}
Section~\ref{sec:intro} claims the rule is violated in none of the eighteen
arms, but Figure~\ref{fig:signature} and Table~\ref{tab:saturation} show the
eleven unhinted arms, so Table~\ref{tab:intervention} gives the other seven.
The conservative form is strictly positive in one and tied in six, which is the
collapse described in Section~\ref{sec:signature}: an intervention that drives
idempotency to 100\% makes the strongest idempotency hazard equal to a
saturated ordering hazard, and the comparison stops discriminating. It is
violated in none. Under the weakest-link form the same seven are positive or
tied, never inverted. Combining the two tables, eleven of the eighteen arms are
strictly positive under the conservative form and none is violated, which is
the claim in Section~\ref{sec:signature}.

\begin{table}[t]
\centering
\caption{The signature rule on the seven intervention arms, the ones not shown
in Figure~\ref{fig:signature}. Rates are per-hazard temporal-pass. Ties arise
where an intervention saturates idempotency, not where the ordering falls
below it.}
\label{tab:intervention}
\begin{tabular}{lrrrrr}
\toprule
Arm & weakest ord. & weakest idem. & strongest idem. & weak-link & conserv. \\
\midrule
Haiku 4.5 + hint   & 84.7\%  & 62.7\%  & 73.7\%  & $+22.0$pp & $+11.0$pp \\
Sonnet 5 + hint    & 100.0\% & 89.5\%  & 100.0\% & $+10.5$pp & $0.0$pp \\
Opus 5 + hint      & 100.0\% & 100.0\% & 100.0\% & $0.0$pp   & $0.0$pp \\
GPT-5.6 Sol + hint & 100.0\% & 100.0\% & 100.0\% & $0.0$pp   & $0.0$pp \\
\addlinespace
Haiku 4.5 + arch   & 100.0\% & 97.4\%  & 100.0\% & $+2.6$pp  & $0.0$pp \\
Sonnet 5 + arch    & 100.0\% & 44.5\%  & 100.0\% & $+55.5$pp & $0.0$pp \\
Opus 5 + arch      & 100.0\% & 100.0\% & 100.0\% & $0.0$pp   & $0.0$pp \\
\bottomrule
\end{tabular}
\end{table}

\paragraph{What each intervention actually did.} Section~\ref{sec:conversion}
reports one arm in which a prompt converted silent failure into crashes. It is
one of seven, and the gap column alone does not say which happened, so
Table~\ref{tab:whatprompts} gives each intervention beside its own baseline on
all three rates. Five arms lower all-in failure with the error rate flat or
falling, which is repair. One, Sonnet 5 with the hazard hint, raises the error
rate by 28.0 points and moves all-in by 7.0, which is relocation. One, Sonnet 5
with the architecture prompt, is worse on every rate than no prompt at all. A
grader that recorded only ``failed'' would show the first six as improvements
of different sizes and could not separate them.

\begin{table}[t]
\centering
\caption{Each intervention arm against its own unhinted baseline. The gap falls
in six of seven; all-in failure, which counts crashes too, diverges from it
wherever the error rate moves, most sharply for Sonnet 5 with the hazard hint.}
\label{tab:whatprompts}
\begin{tabular}{lrrrr}
\toprule
Arm & gap & error & all-in & $\Delta$ all-in \\
\midrule
Haiku 4.5             & 52.2\% & 30.5\% & 67.5\% & --- \\
\quad + hazard hint   & 35.1\% & 29.5\% & 54.5\% & $-13.0$ \\
\quad + architecture  & 0.0\%  & 32.0\% & 32.0\% & $-35.5$ \\
\addlinespace
Sonnet 5              & 48.2\% & 9.0\%  & 51.0\% & --- \\
\quad + hazard hint   & 10.5\% & 37.0\% & 44.0\% & $-7.0$ \\
\quad + architecture  & 54.4\% & 14.0\% & 61.5\% & $+10.5$ \\
\addlinespace
Opus 5                & 16.7\% & 5.0\%  & 21.0\% & --- \\
\quad + hazard hint   & 0.0\%  & 1.5\%  & 1.5\%  & $-19.5$ \\
\quad + architecture  & 0.0\%  & 6.0\%  & 6.0\%  & $-15.0$ \\
\addlinespace
GPT-5.6 Sol           & 7.0\%  & 0.0\%  & 7.0\%  & --- \\
\quad + hazard hint   & 0.0\%  & 0.0\%  & 0.0\%  & $-7.0$ \\
\bottomrule
\end{tabular}
\end{table}

\paragraph{What the check costs.} Grading needs no model calls: it re-executes
a stored artifact locally, seven or eight times per cell. One arm, 40 tasks at
$k=5$, is 200 cells and 1{,}575 verdicts, and re-grades from stored artifacts
in 230 seconds on a single core, about 1.2 seconds per cell. The whole corpus
of 3{,}561 gradeable cells is roughly 70 core-minutes, and the sweep above ran
in about a quarter of an hour on six cores. Generation is the expensive half of
this benchmark and is the half already paid for by anyone running an agent; the
check that separates these models is the cheap half.

\paragraph{Declaration counts.} Late arrival is declared by 38 of 40 tasks and
out-of-order change data by 35, against 40 for each idempotency hazard, so the
two classes' means are taken over slightly different subsets. Recomputed over
the 35 tasks declaring all five, the rule is strictly positive in eleven of
eleven, from $+5.3$ to $+47.1$pp, the minimum being Codestral.

\paragraph{The architecture arms and the guard.} The single exception to the
within-arm split is DeepSeek V4 Pro, where the two rates are 42.2\% and 40.0\%
on 30 unguarded cells. The three architecture arms
are consistent with the within-arm split, carrying guard rates of 97.5\% and
98.9\% in the two that reached a near-zero gap and 33.5\% in the one that did
not. Three arms at three model tiers confound guard rate with capability, so we
do not read them as evidence. We prefer the within-arm split for the same
reason to an across-arm correlation of aggregates, which mixes arms differing
in far more than guard rate and gives only $r = 0.64$; nor does the guard
generalise beyond its own hazard, that correlation falling to $r = -0.17$
($\rho = -0.43$) once duplicate delivery is excluded.

\paragraph{The engine confound, and the sign tests.} The signature could in
principle be a property of this dialect rather than of model knowledge: DuckDB
window functions and \texttt{ORDER BY} make event-time ordering natural to
express, while there is no upsert idiom the models reach for. We cannot
separate these with one engine. The human arm is a partial answer, since the
participants wrote SQL on the same engine, against the same hazards, and
inverted the ordering anyway; seven to nine cells make that suggestive and not
conclusive. A second engine is the single most valuable extension of this work.
Both sign tests in the paper are one-sided. The signature test needs the
right null: the statistic compares the weaker of two ordering hazards against
the weakest of three idempotency hazards, so under exchangeability of the five
it succeeds whenever the globally weakest hazard falls in the three-group, with
probability $3/5$ and not $1/2$. Against that null, eleven of eleven gives
$p = 3.6\times10^{-3}$. The
guard test gives $P(X \geq 12 \mid n=13, p=1/2) = 14/8192 = 1.7\times10^{-3}$.
Neither is a claim we lean on, for the independence reasons given in the body.

\paragraph{Watermarks and the late-arrival hazard.} Spark's guide documents
that data beyond a watermark threshold is dropped, so a pipeline that discards
late events could be said to follow it. Our execution contract names no
watermark and no task declares a lateness bound; what is required is
convergence to the batch recomputation, and the deep-lateness component of the
schedule delays 5\% of eligible events to the final batch, beyond any
threshold short of the whole run. The behaviour is graded as a failure by
design rather than by accident, and the negative control
\emph{broken\_watermark\_drop.sql} pins that decision: it implements the
documented drop, passes snapshot evaluation, and is declared to fail. A
benchmark for a system that opts into bounded state would score it the other
way, and would need to say so in the task.

\paragraph{Withdrawn claims.} The three are a non-monotonicity in tier
ordering, that hints do not help, and that hints induce fragility. Each is why
this work pre-registers its analysis and lets code, not prose, decide when an
interval is too wide to support a claim.

\paragraph{Human baseline, per-cell detail.} Our registered expectation of a
5--15\% human gap was wrong, over-predicting their competence. Under the same per-hazard rule as
the model arms, counting each hazard over the cells where that hazard itself
did not error, the humans pass late arrival 3/7, out-of-order change data 3/7,
duplicate delivery 6/9, crash-retry 6/9 and backfill 5/9. Under the stricter
gap denominator, which drops a cell if any declared hazard errored, the counts
are 3/7, 3/7, 6/7, 6/7 and 4/7; the body uses the former so that the human and
model figures share an error-exclusion rule. They do not share an aggregation
rule, and Section~\ref{sec:humanarm} says which each uses. All four participants declared a primary key unprompted on at least one
task;
where that fired on redelivery it produced a loud crash rather than silent
corruption, the less costly of the two failures. The tasks attempted were
\emph{customer\_state\_modify}, \emph{se\_net\_revenue},
\emph{h\_rolling7\_repair}, \emph{h\_top3\_daily} and
\emph{t2\_device\_state}; the model figures in
Section~\ref{sec:humanarm} are recomputed over exactly this set. A different
task sample could move these counts substantially.

\section{The agent, the prompts, and the human arm}
\label{app:agent}

\paragraph{The loop.} Each cell is a fixed-turn loop of at most three turns.
Turn one sends the system prompt below, the task, its source and target
schemas, and the snapshot test. The agent returns one DuckDB script. The
harness executes that script \emph{on the snapshot path only} and either ends
the loop on a pass, returns the DuckDB error verbatim on an execution failure,
or returns a typed diff summary with up to three example rows when the target
does not match. No feedback names a hazard, mentions a delivery schedule, or
reports a hazard verdict, and the word-boundary leak scan applies to it. Only
the agent's final script is graded, on the snapshot path and on every schedule.
Sample 0 is drawn at temperature 0 and samples 1--4 at 1.0 for models that
accept sampling parameters; Opus 5, Sonnet 5 and the GPT-5.6 family reject
them, so the parameter is omitted and those arms vary on their own. Output is
capped at 8{,}000 tokens per turn, the cap that truncated fourteen cells.

\paragraph{What the loop does to snapshot-pass.} The loop repairs against
exactly the check the incumbent metric applies, so snapshot-pass in this paper
is a best case for that metric rather than a neutral measurement. Pooled over
the 2{,}200 unhinted cells, 79.5\% pass the snapshot on the first turn. Per arm,
the first-turn rate and the final rate are:

\begin{table}[h]
\centering
\caption{Snapshot-pass before and after the repair loop, means over tasks. The
loop is worth up to 32.5 points, and the arms it helps most are the ones the
incumbent metric would otherwise separate.}
\label{tab:repairloop}
\begin{tabular}{lrrr}
\toprule
Arm & turn 1 & final & lift \\
\midrule
Haiku 4.5         & 82.5\%  & 98.9\%  & $+16.4$ \\
Sonnet 5          & 95.5\%  & 100.0\% & $+4.5$ \\
Opus 5            & 94.5\%  & 100.0\% & $+5.5$ \\
GPT-5.6 Sol       & 100.0\% & 100.0\% & $0.0$ \\
\addlinespace
Codestral         & 64.0\%  & 86.0\%  & $+22.0$ \\
Qwen3-Coder-Next  & 68.0\%  & 99.2\%  & $+31.2$ \\
Gemma 4 26B       & 77.5\%  & 99.5\%  & $+22.0$ \\
DeepSeek V4 Flash & 70.0\%  & 97.9\%  & $+27.9$ \\
DeepSeek V4 Pro   & 67.5\%  & 100.0\% & $+32.5$ \\
MiniMax M3        & 68.0\%  & 99.5\%  & $+31.5$ \\
Kimi K3           & 86.5\%  & 99.5\%  & $+13.0$ \\
\bottomrule
\end{tabular}
\end{table}

Without the loop the incumbent metric would separate these models somewhat,
across 64.0--100.0\%. It still would not resolve them the way temporal
evaluation does, across 7.0--79.2\%, and the arms remain in a different order.
We report the post-loop figure as the headline because it is the condition the
agents actually ran under.

\paragraph{The system prompt, identical in all three arms.}

{\small
\begin{verbatim}
You are a data engineer. You write incremental pipelines in DuckDB SQL.

EXECUTION CONTRACT - read carefully, it defines what your script must do:
- Your script is ONE DuckDB SQL script. It is run ONCE PER DELIVERY BATCH
  against a database that PERSISTS between runs.
- Before each run the harness creates `<source>_batch`, containing ONLY the
  current batch's arrivals, with columns:
    event_id BIGINT, ts_event TIMESTAMP, <payload columns...>,
    _arrival_seq BIGINT, _batch_id BIGINT
- There is NO cumulative table of past arrivals. If your pipeline needs
  history, it must persist that history itself.
- Your script must create and maintain the target table. Use
  CREATE TABLE IF NOT EXISTS - the script runs many times.
- Any other tables you create also persist between runs; that is your state.
- `ts_event` is event time (logical). `_arrival_seq` and `_batch_id` are
  arrival metadata (physical).

Respond with the SQL script and nothing else, in a single ```sql code block.
No prose outside the block.
\end{verbatim}
}

The user turn appends the task name and description, the source and target
column lists, the target key, the existing pipeline for \textsc{modify} and
\textsc{repair} tasks, and this closing paragraph:

{\small
\begin{verbatim}
## Test
Your pipeline is checked by loading the complete event log as a single
batch, running your script once, and comparing `<target>` to the
expected result.
\end{verbatim}
}

\paragraph{The hazard hint, appended in the four hint arms.}

{\small
\begin{verbatim}
NOTE: the pipeline will be executed under adversarial delivery schedules of
the same logical data: late-arriving events, out-of-order updates for the same
key, duplicate delivery, crash-and-retry mid-script, and backfill of a
historical partition. The final state must equal a batch recomputation over
the complete event log under every one of them.
\end{verbatim}
}

\paragraph{The architecture prompt, appended in the three architecture arms.}
This is the prompt Section~\ref{sec:displacement} holds responsible: it
specifies deduplication against stored history and says nothing about
duplicates arriving together.

{\small
\begin{verbatim}
ARCHITECTURE GUIDANCE: structure your script as two stages.

  1. STAGE: append this batch's rows into a persistent staging table that
     holds the full history of what has been delivered, keyed on event_id,
     inserting a given event_id only if it is not already present.
  2. DERIVE: compute the target table as a pure function of that staging
     table, rather than by modifying the target in place.
\end{verbatim}
}

\paragraph{How the human tasks were chosen, and how the arm was run.} The five
tasks were selected before any submission arrived, by a deterministic greedy
rule that prefers a task adding an uncovered hazard, then an uncovered task
mode, then an uncovered difficulty band, then whichever balances the count of
snapshot-hard tasks, with ties broken on a fixed seed rather than on process
hash. The selection covers all six hazards, three task modes and three
difficulty bands, and is recorded with its rationale. Participants were
practising data engineers known to the author, recruited directly and
unpaid; the packet carried a consent page, the identical unhinted prompt,
and a record sheet on which each participant logged minutes per task and could
decline a task. Four of the five invited returned work, giving nineteen cells
over five tasks with one task declined. They worked untimed on their own
machines with no repair loop, which is the main reason their snapshot-pass
rate of 68.4\% sits below every model arm. Neither the no-prior-knowledge
declaration nor the no-assistant request was enforced; we inspected the
submitted scripts for assistant boilerplate and found none, which is the only
evidence available and it is weak.

\paragraph{Three questions this design invites.} The expected result is a
\emph{batch recomputation}, not the gold pipeline run once: the oracle rebuilds
the target from the complete logical log with an independent implementation, so
a gold that was itself wrong under a schedule could not certify itself.
Backfill both redelivers processed events and releases withheld ones, so it is
a mixed hazard classed with idempotency; the weakest idempotency hazard is
duplicate delivery in nine of the eleven unhinted arms, so the signature does
not rest on backfill's classification. And the two benign controls are graded
by the same oracle as the hazards, which is what makes the in-order stream
result interpretable as ordinary incrementality failure.

\end{document}